\documentclass[journal]{IEEEtran}
\IEEEoverridecommandlockouts

\usepackage{include}

\begin{document}

\title{Reinforcement Learning Compensated Model Predictive Control for Off-road Driving on Unknown Deformable Terrain}

\author{
    {Prakhar Gupta, Jonathon M. Smereka, Yunyi Jia}%
    \thanks{Prakhar Gupta ({\tt\footnotesize prakhag@clemson.edu}) and Yunyi Jia ({\tt\footnotesize yunyij@clemson.edu} are with the Department of Automotive Engineering, Clemson University, Greenville, SC 29607, USA.}%
    \thanks{Jonathon M. Smereka ({\tt\footnotesize jonathon.m.smereka.civ@army.mil}) is with the Ground Vehicle Systems Center, U.S. Army Combat Capabilities Development Command, Warren, MI 48397 USA.}%
    \thanks{DISTRIBUTION STATEMENT A. Approved for public release; distribution is unlimited. OPSEC \# 8779}%
}

\maketitle

\begin{abstract}
This study presents an Actor-Critic reinforcement learning Compensated Model Predictive Controller (\texttt{AC\textsuperscript{2}MPC}) designed for high-speed, off-road autonomous driving on deformable terrains. Addressing the difficulty of modelling unknown tire-terrain interaction and ensuring real-time control feasibility and performance, this framework integrates deep reinforcement learning with a model predictive controller to manage unmodeled nonlinear dynamics. 
We evaluate the controller framework over constant and varying velocity profiles using high-fidelity simulator Project Chrono. Our findings demonstrate that our controller statistically outperforms standalone model-based and learning-based controllers over three unknown terrains that represent sandy deformable track, sandy and rocky track and cohesive clay like deformable soil track. Despite varied and previously unseen terrain characteristics, this framework generalized well enough to track longitudinal reference speeds with the least error.
Furthermore, this framework required significantly less training data compared to purely learning based controller, converging in fewer steps while delivering better performance. Even when under-trained, this controller outperformed the standalone controllers, highlighting its potential for safer and more efficient real-world deployment.
\end{abstract}


% FOR ARXIV submission - add IEEE notice
\makeatletter
\def\ps@IEEEtitlepagestyle{ % Standard Letter Document Class package 
    \def\@oddfoot{\mycopyrightnotice}
}
\def\mycopyrightnotice{
    {\footnotesize This work has been submitted to the IEEE for possible publication. Copyright may be transferred without notice, after which this version may no longer be accessible.\hfill}
}

%%%%%%%%%%%%%%%%%%%%
%%% Introduction %%%
%%%%%%%%%%%%%%%%%%%%
\section{Introduction}

% High speed and dynamic off-road autonomous driving applications for manned and unmanned vehicles have been the focus of research to advance control, perception and decision making algorithms. From a control perspective, the challenge is to formulate a vehicle controller that can perform on the uneven, uncertain and deformable terrain. This is not always realizable  because the tire-terrain interaction is difficult to model with high confidence. In cases where a fully representative dynamics can be modelled, solving the complex formulation for control in real-time is intractable.
% Model-based control has the benefit of theoretical guarantees for safety and stability. But the relatively simple real-time non-linear model-based controllers \cite{KinematicBicycleModel2015} that perform well on road-like surfaces fail to perform because a terrain model mismatch arises and lends to inaccurate control generation. Purely learning based control can be exploited to learn controls for a given driving scenario, but lacks the ability to assure safety unless augmented with explicit safety techniques like barrier functions, model-based filters etc. This is where hybrid approaches can contribute to improve controls by leveraging the advantages of model-based and learning-based controls.

High-speed and dynamic off-road autonomous driving for both manned and unmanned vehicles has been a focal point of research aimed at advancing control, perception, and decision-making algorithms. From a control perspective, the primary challenge lies in designing a vehicle controller capable of operating on uneven, uncertain, and deformable terrain. Achieving this is difficult because modeling tire-terrain interactions with high confidence is inherently challenging. Even when a fully representative dynamics model is developed, solving the complex control formulation in real-time often remains infeasible.

Model-based control offers theoretical guarantees for safety and stability. However, relatively simple real-time nonlinear model-based controllers, such as the non-linear bicycle model, which perform well on road-like surfaces \cite{KinematicBicycleModel2015}, can fail in off-road conditions due to terrain model mismatches that lead to inaccurate control input generation. Purely learning-based control methods can learn control policies for specific driving scenarios but lack inherent safety assurances unless supplemented with explicit safety techniques like barrier functions or model-based filters \cite{safeRL}. This is where hybrid approaches come into play, combining the strengths of both model-based and learning-based controls to enhance performance and safety.

Existing work on hybrid controls discussed below predominantly focuses on on-road surfaces or quadrotors, leaving a gap in hybrid reinforcement learning controls for vehicles on highly deformable terrains. Moreover, none of these works below explore utilizing model predictive control and reinforcement learning together for data efficient controls to address high-speed, off-road driving on unknown deformable terrain.
\begin{figure*}[]
    \centering
    % \includesvg[width=\linewidth]{images/architecture.svg}
    \includegraphics[width=\linewidth]{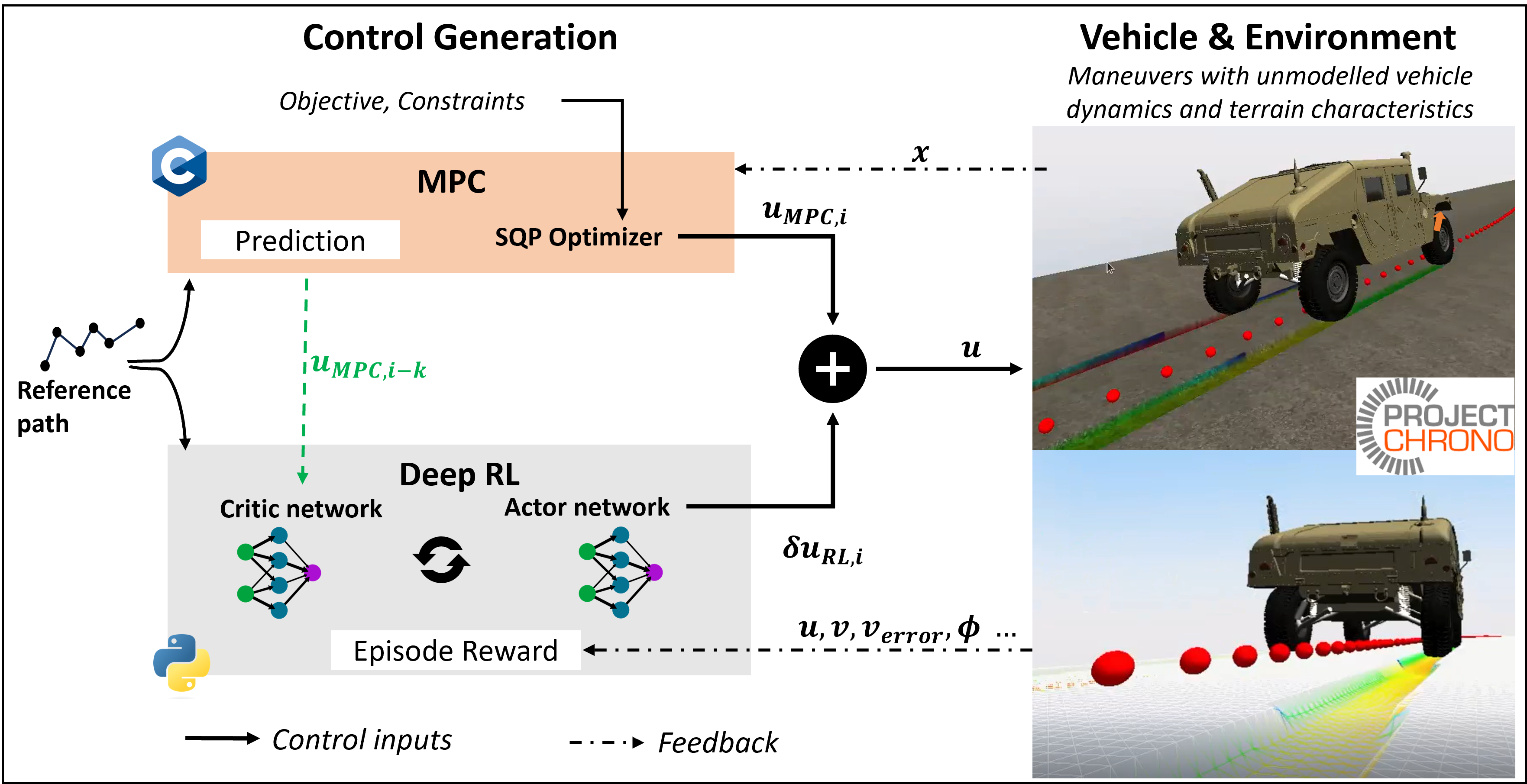}
    \caption{\texttt{AC\textsuperscript{2}MPC} controller training and simulation framework.}
    \label{fig:architecture}
\end{figure*}
\subsection{Relevant Work} \label{sec:relevant}
A rigorously studied set of approaches to handle uncertain but characterizable dynamics is to augment the non-linear MPC formulations using robust \cite{bemporad2007robust} and stochastic schemes \cite{Mesbah}, \cite{saltik2018outlook}. These schemes introduce additional optimizations or optimization elements such as chance constraints, which require significant effort to re-formulate for solution in real-time applications such as drone flights, scaled vehicle racing etc. Since characterization of the disturbances or their probability distributions is not straight-forward, these methods are cumbersome to implement. Adaptive methods that exploit the robust control and learning ideas have been used in \cite{bujarbaruah2019adaptive, rosolia2017learning, sinha2022adaptive}. However, these approaches still require a known uncertainty bound to perform, or a repetitive task so that the learning can converge through set membership methods. In many cases they are also limited to only linear systems, such as in \cite{strelnikova2024adaptive}. In case of an unknown terrain with uncharacterized uncertainty bounds, these approaches can be difficult to implement.

Learning based optimal control approaches in \cite{hewing2020learning}, \cite{fisac2018general} use gaussian process (GP) regression to learn the unknown dynamics and propagate their uncertainty bounds forward in time. Here, the unknown dynamics is treated as the model error that is added to the nominal dynamics and the MPC optimization problem is solved using reachable sets. This can update the confidence bounds on the go, but the time complexity of GPs scales poorly with number of states and sparse approximations were required to arrive at a tractable solution times. 
Sampling based control and optimization that resemble the MPC framework \cite{williams2017model, bootsRL, 10190183} have been explored lately for off-road driving. These sometimes avoid running a gradient descent through a large number of control space sampling and brute-forcing to calculate the least cost roll-out trajectory. However, a downside of these approaches is that they don't always support state constraints for safety and scale poorly with the size of control space without parallelization.

Another set of approaches is where a learning based controller is primarily employed used during deployment. Here, the controller is either trained to imitate aspects of an expert, i.e., the model-based controller itself, or learn a policy through data and observations and back-propagation techniques. 
A convolutional network was trained to control the vehicle steering and throttle in \cite{bojarski2016end} for driving on a path using vision sensors.
In \cite{salzmann2023real} a deep neural network is trained in to imitate a non linear program, but it requires linearization and feasibility checks for control synthesis. 
In \cite{spielberg2021neural} a neural network is trained to learn to drive autonomously on an unknown friction surface. This approach performs well for their use case, but there is little comment on the performance when the states go  out-of-distribution with respect to the training data. 
\cite{ahn2023model} takes a step towards addressing out of distribution behavior for safety through positive invariance, for a control affine dynamic system.
\cite{sacks2023deep} learns the inner optimization loop of an MPC to control a quad-rotor platform. 
Deep RL control has demonstrated promise to learn complex continuous control policies from data since the work in \cite{lillicrap2015continuous}. These approaches use neural networks to model the controller and update the network through reward functions.
Others like \cite{zhang2019safe}, \cite{maddalena2020neural} also utilize neural networks to learn from data and selected scenarios. 
Residual Policy Learning (RPL) \cite{silver2018residual}, \cite{kerbel2023adaptive} exploits the idea of building up on top of an existing controller that performs a task in an acceptable manner to some extent. The neural network is first trained to imitate the base policy and then exposed to more training data so that the control policy learns to generalize for model mismatch or uncertainty. These controllers were shown to outperform all the base policies they were trained on.
However, all of these approaches require significant amount of data to perform well and pose the risk of unsafe control generation for out-of-distribution states.

In general, the approaches where a learning based controller is the primary controller during safety critical deployment, additional model-based methods are employed to certify safety. Deploying model based controllers alongside learning methods allows improvement in control performance and safety. Among others, this has been explored using optimal controllers with Bayesian optimization techniques to find optimal parameters \cite{sorourifar2021data} and deep neural networks of varying characteristics. 

Approaches that specifically combine aspects of MPC and RL paradigms for control improvement include learning the parameters of the model based controller. In \cite{zanon2020safe}, a robust MPC is parameterized and these parameters learnt through reinforcement learning. In these approaches, a system model and controller formulation of significant fidelity with tunable parameters is inevitably required and difficult for an off-road driving application.

There have been efforts that utilize reinforcement learning to arrive at optimal cost functions, weights and values for the model-based controller. The cost function value of the optimal controller is updated using reinforcement learning's value estimation in \cite{bhardwaj2020blending} and the terminal cost of the MPC is approximated through reinforcement learning in \cite{arroyo2022reinforced}. Through learning stage and terminal cost values, these approaches attempt to indirectly address mismatches in modelling and accommodate long term planning respectively. More recently in \cite{scarramuzza}, a differentiable MPC is embedded in the last layer of the deep RL network. The choice of cost function weights is learnt here, to arrive at a more generalized control policy altogether. The limitation of the differentiable MPC here is its inability to handle state constraints.
In \cite{zarrouki2024safe}, the agent learns to tune the optimization controller penalty weights, while sampling from only a set of weights that lead to safe actions.
However, these approaches are computationally very expensive as they require either multiple passes over the network or multiple stages of optimization.

Deriving from the ideas of imitation learning, in \cite{ETh_RL+Model}  the agent learns from the reference trajectories of the legged robot generated by the MPC at the beginning of each episode, and only the agent is used to generate control during deployment for a more versatile motion.
\cite{kim2024reinforcement} uses RL to learn a more general lane change planning using the MPC as the expert policy and in \cite{bootsRL}, the system dynamics for the model predictive path integral controller is learnt through RL. The use of a neural network to sample the dynamic trajectories increases the efficiency of their framework, while allowing a more generalized system model. 
\cite{delft}, where a hierarchical approach is adopted where MPC is used as long term planner and the agent learns to manipulate the planned inputs at a higher frequency. The framework is tested for freeway traffic control applications, where long term behavior learning can augment MPC's short term focus. Since this is a hierarchical approach, the learning based controller is effectively the primary control generator. 
Since the the learning agent ends up effectively being the primary controller in these approaches, they do not promise confidence in unseen scenarios.

\subsection{Contributions}
In this work, we propose a non-hierarchical learning-based compensation framework integrated with a model-based control system. Specifically, we employ a parallel actor-critic deep reinforcement learning controller (AC) that learns to manipulate the control inputs from the model predictive controller (MPC) at every step, as illustrated in Fig.~\ref{fig:architecture}. This approach aims to address the unmodeled nonlinear dynamics arising from terrain characteristics, allowing the compensatory policy to focus solely on the additional control inputs necessary to account for these unmodeled dynamics. The framework leverages the online solution of the optimal control problem to develop a compensation controller policy that enhances longitudinal tracking performance on deformable terrain.
The contributions of this work can be summarised as follows:
\begin{itemize}
    \item We develop a parallel compensation architecture, \texttt{AC\textsuperscript{2}MPC}, using deep reinforcement learning to improve the longitudinal performance of a model-based optimization controller for off-road driving.
    % \item The framework allows building in cooperation and awareness of compensatory learning-based controller, which ensures that MPC is the primary controller and yields smooth overall control. This way the agent learns to intervene only in the operating regions where primary controller cannot achieve target tracking.
    \item By not requiring the learning agent to develop a complete policy from scratch, we demonstrate that the data requirements for training the framework are reduced.
    \item For online learning and deployment, we show that this framework outperforms both standalone AC and MPC controllers, even when under-trained.
    \item The framework is validated using a high-fidelity off-road driving simulator with deformable terrain physics and tested in environments not encountered during the agent's training.
\end{itemize}

The rest of this paper will discuss how MPC and AC are utilized in a parallel non-heirarchical fashion and its evaluation for longitudinal performance on deformable terrain in simulation.

\section{Compensated Control Framework} 
\subsection{Model Predictive Controller}\label{sec:mpc}
As the dynamic nature of the desired motion increases, minimizing position overshoots and respecting safety constraints during trajectory tracking becomes essential. For autonomous driving off the road, optimization based and predictive controllers have been in most explored lately \cite{williams2017model, 10190183}. For this reason, we choose MPC as the baseline model-based controller to predict and optimize future actions for a minimum tracking error and smooth controls with explicit safety constraint handling on deformable terrain. The formulation for this predictive optimization is detailed in Eq. \eqref{eq:ll-mpc}.
\begin{mini!}
    {
        \mathbf{u}, \, \mathbf{x}}{
        \left\Vert x_N - \rho_{N} \right\Vert^2_{P} + \ldots
        %\qquad\qquad\qquad\qquad\qquad
    }{\label{eq:ll-mpc}}{}
    \breakObjective{\sum_{i=0}^{N-1} \left( 
            \left\Vert x_i - \rho_{i} \right\Vert^2_Q + \left\Vert u_{i} - \mu_{i} \right\Vert^2_R
        \right)}
    \addConstraint{\label{eq:llmpc_dynamics}
        x_{i+1} = f(x_i, u_i)
    }{}%\forall \ i = 0, \ldots, N-1}
    \addConstraint{\label{eq:llmpc_outputs}
        y_i = a_l = \frac{v_i ^ 2}{L} \tan(\theta_i)
    }{}
    \addConstraint{\label{eq:llmpc_c1}
        0 \leq v_i
    }{}
    \addConstraint{\label{eq:llmpc_c2}
        \underline{a}_l \leq a_l \leq \overline{a}_l
    }{}
    \addConstraint{\label{eq:llmpc_c3}
        \underline{a_{mpc}} \leq a_i \leq \overline{a_{mpc}}
    }{}
    \addConstraint{\label{eq:llmpc_c4}
        \underline{\theta} \leq \theta_i \leq \overline{\theta}
    }{}
    \addConstraint{\label{eq:llmpc_c5}
        \underline{\omega} \leq \omega_i \leq \overline{\omega}
    }{}
\end{mini!}
where,
$ x_i $ is the vehicle state vector at any given time $i$;
$ u_{i} = \left[ a, \omega \right] $ is the control input vector of forward acceleration and rate of change in steering angle; $\mathbf{x}$ and $\mathbf{u}$ are the respective state and control trajectories over $N$ stages that enter as decision variables for the optimization; $\rho_i$ and $\mu_i$ are the respective reference states and controls at each control stage $i$ corresponding to the trajectory planned ahead in time;
$ Q \succeq 0 $ is the symmetric positive semi-definite state penalty matrix; 
$P_f \succeq 0 $ is the symmetric positive semi-definite terminal state penalty matrix; 
and $ R \succ 0 $ is the symmetric positive definite input penalty matrix.

For the system model $f(x_{i}, u_{i})$ of Eq. \eqref{eq:llmpc_dynamics} we choose a kinematic bicycle model as in \cite{KinematicBicycleModel2015}, where the use of these kinematic bicycle models in low level vehicle control was found to have acceptable performance for on-road driving. This model leads to an optimizer formulation that is non-linear yet simple enough to solve in real-time. 
\begin{subequations} \label{CTbicycle1}
    \begin{align}
        \Dot{s}_x &= v \cos{(\phi + \beta)} \\
        \Dot{s}_y &= v \sin{(\phi + \beta)} \\
        \Dot{\phi} &= \frac{v}{L} \tan{(\beta)} \\
        \Dot{\theta} &= \omega \\
        \Dot{v} &= a_{mpc} \\
        \beta &= {\tan}^{-1}(\frac{L_r}{L} \cdot \tan{\theta})
    \end{align}
\end{subequations}
where, the states chosen for $ x_i = \left[ s_x, s_y, \phi, \theta, \beta \right]$ are, respectively, the: $(x,y)$ coordinate positions in the map frame, heading in the map frame, steering angle, forward vehicle speed and body slip angle. This formulation choice has been exploited in our previous physical vehicle on-road experiments with success \cite{Gupta2023} and thus lends a relevant baseline for this study.
It should be noted that the non-linear kinematic bicycle model that ignores the body slip angle approximation as in \cite{lynch2017modern} yields similar performance. This is because they both represent the same assumptions of kinematic behavior.

We impose bounded state constraints on velocity and steering angle to prevent the vehicle from reversing and to match realistic steering column limits - where the minimum and maximum steering angles $\underline{\theta}$, $\overline{\theta}$ limit the steering magnitude. Additionally, bounded constraints are imposed on the control inputs to prevent harsh vehicle acceleration and prevent rapid changes in steering angle - where  $\underline{a_{mpc}}$, $\overline{a_{mpc}}$ are the minimum and maximum longitudinal accelerations to limit the acceleration magnitude within \unit[5]{m/s$^2$}; and  $\underline{\omega}$, $\overline{\omega}$ limit the minimum and maximum steering rate of change. Finally, safety constraints are imposed on the magnitude of the lateral vehicle acceleration $a_l$. These are enumerated in Table \ref{tab:controlParams}.

The \texttt{acados} optimization library \cite{acados} is used to generate a \texttt{C}-based SQP and SQP-RTI solvers for optimization \eqref{eq:ll-mpc} utilizing a 4\textsuperscript{th}-order implicit Runge-Kutta integrator and direct multiple-shooting method. Commands from this feedback controller are sent to the vehicle simulator used in this study. It should be noted that throttle control inputs for all controllers are normalized such that unity which corresponds to \unit[5]{m/s$^2$}.

\subsubsection*{\textbf{Online Reference Generation for MPC}}
\begin{figure}[]
    \centering
    % \includesvg[width=\linewidth]{images/refGen.svg}
    \includegraphics[width=\linewidth]{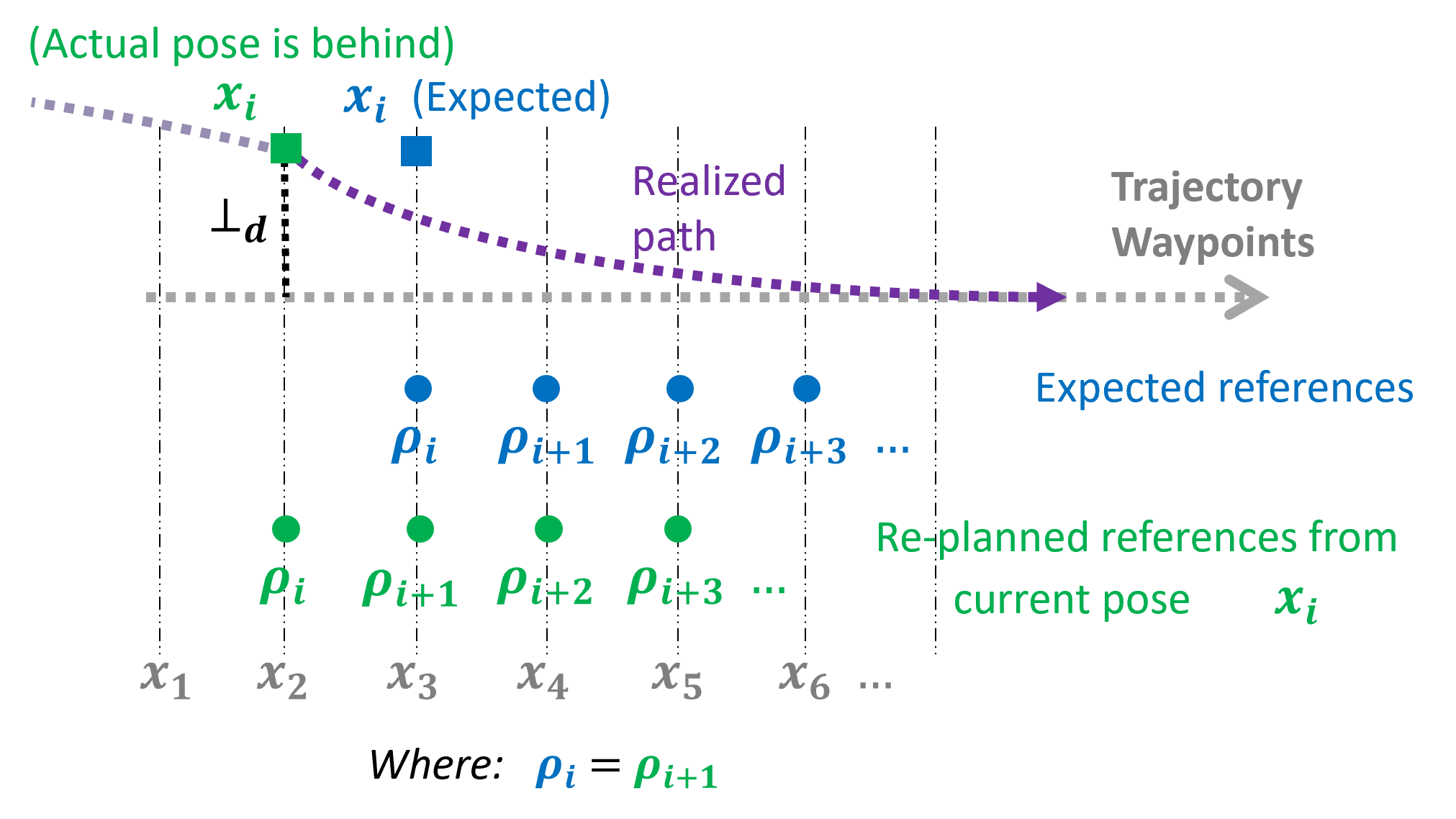}
    \caption{Online reference generation at time-step $i$: In the case where not enough progress is made along the trajectory by $i$, re-planned green references from actual $x_i$ are passed, instead of passing the blue references which represent the references if expected progress were made.}
    \label{fig:refGen}
\end{figure}

Offline generated references can lead to solution infeasible or jerky controls from the optimizer when progress made along the trajectory is not as predicted in the previous time-step. To ensure smoothness and feasibility of control optimization during run-time, especially on a real vehicle as in our previous experiments 
\cite{Gupta2023}, the references are computed online in closed-loop. This is based on the current state ($x_i$) and shortest distance to desired trajectory ($\perp_d$), as shown in Fig \ref{fig:refGen}. This prevents the MPC attempting to achieve the target trajectories through unreasonable control inputs. However, this is a choice that yields a framework that is not aware of time along the trajectory, and only the position. We consider this to be a good choice for off-road controls because position matters significantly when driving in obstacle ridden or around regions of intraversibility.

This controller is used to simulate velocity tracking control for a vehicle with three levels of increasing model mismatch as shown in Fig.~\ref{fig:rigid-def}. 

\subsection{Actor-Critic Reinforcement Learning Formulation} \label{sec:rl}
Unlike the MPC formulation in Section \ref{sec:mpc} that does not have the ability to handle unknown and mismatched dynamics, reinforcement learning based controls can learn to carry out control tasks for a given unnknown environment through the correct choice of reward functions. Through an explore-exploit approach, the agent's action policy gradually converges to one that yields the best value for its actions.
In vehicle controls, this was made possible through implementation of reinforcement learning algorithms for continuous action paradigm and initially done in \cite{lillicrap2015continuous} using policy optimization to update neural networks. A typical policy performance gradient equation is given by

\begin{subequations} \label{eq:rl-pg}
    \begin{align}
        \theta_{k+1} &= \theta_k + \alpha \nabla_{\theta} J(\pi_{\theta_k})\\
        \nabla_{\theta} J(\pi_{\theta_k}) &= \displaystyle \mathop{\mathbb{E}}_{\tau\in \pi_{\theta}}[\sum_{t=0}^{T} \nabla_{\theta} \log \pi_{\theta}(a_t | s_t) \cdot R(\tau) ]
    \end{align}
\end{subequations}

where, the neural networks then represent the control policy $\pi_{\theta}$ with their parameters $\theta_k$ and timestep $k$. $\tau$ is the trajectory rollout under the given policy and $R(\tau)$ is the cumulative reward from the trajectory. In this manner, the trajectory samples generated through interaction with the environment can be used to conduct probabilistic gradient updates.

For generality of control performance across different dynamics maneuver and trajectories, this study leverages the Proximal Policy Optimization (PPO) actor-critic reinforcement learning algorithm \cite{schulman2017proximal}, which is known for its stability and robustness to hyper-parameters. Here, instead of running a policy gradient, the we update the parameters $\theta$ in a constrained clipping manner using KL divergence criteria.
\begin{subequations} \label{eq:rl-ppo}
    \begin{align}
        \theta_{k+1} &= \arg \max_{\theta} \mathcal{L}(\theta_k, \theta)\\
        & \text{s.t.} \quad \overline{D}_{KL}(\theta || \theta_{k}) \leq \delta
    \end{align}
\end{subequations}

where $\mathcal{L}$ is the surrogate advantage function that represents how the new policy performs in comparison to old policy, and $\overline{D}_{KL}$ is the mean KL divergence between the policies.
A continuous action and observation space implementation is chosen to represent the continuous nature of vehicle states and controls. 
To train the RL agent for throttle control, the observation space consists of current vehicle speed, target speed, action history taken and speed tracking error history. The action space consists of just the throttle inputs. The details of hyper-parameters used and action bounds are provided in Table \ref{tab:controlParams}. The neural network architecture chosen for the policy and the target networks is a small one, with the number of neurons in each hidden layer being $\{8,32,16,8\}$ with \texttt{ReLu} activation units.
The following reward function $r_1(i)$ is chosen for the agent.

\begin{subequations} \label{eq:r1}
    \begin{align}
        r_1(i) &= W_{11} \cdot \frac{1}{1+ \left| v_{err} \right|} - W_{12} \cdot \frac{\sigma_{a_{rl}}}{N_{\sigma}} - W_{13}\\
            & \text{s.t. } W_{13}=\begin{cases}
            1, & \text{if $v < 0$}\\
            0, & \text{otherwise}.
          \end{cases}
    \end{align}
\end{subequations}
Here, we penalize the speed tracking error  $v_{err}$ and negative velocities while encouraging a low standard deviation $\sigma_{a_{rl}}$ over the observed action history span of $h_a$ to generate smooth control actions. These components are all normalized to unity with the help of appropriate normalizer $N_{\sigma}$ to improve learning. The corresponding weight terms $W_{1k}$ are non-negative real numbers.
Since the focus of this study is longitudinal control on deformable terrain, a simple proportional controller is used for steering controls along with the RL throttle input.

The algorithm is implemented using the well developed \texttt{stable-baselines3} package \cite{stable-baselines3} that allows multi-processing to speed up the learning process, and the Adam optimizer \cite{kingma2014adam} is used for the running the policy gradient on the networks.

\begin{table}[]
    \centering
    \caption{Control and learning parameters.}
    \begin{tabular}{r|lll}
    \toprule
        Controller & Symbol & Description & Value \\
    \midrule
    \midrule
        All & $f_c$ & control frequency &  10 [hz] \\
            & $T_s$ & simulation time-step &  $3e^{-3}$ [sec] \\
            &  $\underline{a}, \overline{a}$ & actuation limits & $[-1,1]$ \\
    \midrule
         MPC & $a_l$ & lateral acc bound & $ \pm 1.5 [m/s^2]$ \\
        (SQP)     & $N$ & control stages &  10 \\
                & $T_h$ & horizon time &  5 [sec] \\
                & $L$ & wheelbase &  2.75 [m] \\
                & $L_r$ & CG to rear axle &  1.75 [m] \\
                & $\underline{\theta}, \overline{\theta}$ & steering bound & $\pm 0.57 [rad]$ \\
                &  $\underline{\omega}, \overline{\omega}$ & steer rate bound & $\pm 0.05 [rad/s]$ \\
                &  $\underline{a_{mpc}}, \overline{a_{mpc}}$ & action bounds & $[-1,1]$ \\
    \midrule
        AC  & $\underline{a_{rl}}, \overline{a_{rl}}$ & action bounds & $[-1,1]$\\
         (PPO)   &  & observation size & 12 \\
                &  & action size & 1 \\
                & $h_a$ & action history size & 10 \\
                & & learning rate & 0.01 \\
                & & clip factor & 0.2 \\
                & $b_{size}$ & batch size & 50 \\
                &  & steps per epoch & 300 \\
                
    \midrule
        AC\textsuperscript{2}MPC  & $\underline{a_{rl}}, \overline{a_{rl}}$ & RL action bounds & $[-1,1]$ \\  
        (SQP+PPO)    &  & observation size & 32 \\
                & $h_a$ & action history size & 10 \\

    \bottomrule
    \end{tabular}
    \label{tab:controlParams}
\end{table}

\subsection{Ensemble Controller Formulation: \textbf{AC\textsuperscript{2}MPC}}
MPC described in Section \ref{sec:mpc} is adequate for precise tracking when the dynamics of the vehicle matches the dynamics modelled in Eq. \eqref{eq:llmpc_dynamics}. But as shown in Fig.~\ref{fig:rigid-def}, it fails to perform in the event of mismatch.
The deep RL in Section \ref{sec:rl} can learn to perform even when the system dynamics is not modelled through exploration with policy updates, but needs a lot of data to learn a control policy from scratch. It lends to low confidence in regions of operation that lie outside of the training data.
The compensated controller framework attempts to leverage and combine the advantages of these two. We consider the actual high fidelity dynamics to be a combination of the modelled non-linear bicycle model and some unknown dynamics $ f_u(x_i,u_i)$ as shown in Eq. \eqref{eq:fu}. We exploit the idea that a learning based compensatory controller can be added to the baseline MPC closed-loop framework that learns to manipulate the control inputs of the MPC and account for the unmodeled dynamics $f_u$ it observes during training. This leads to the expectation that the agent would need lesser data to learn a compensatory control policy than the standalone control policy as in Section \ref{sec:rl}, and also lead to a control framework that relies primarily on the model based controller. 
% We formulate the compensated controller \texttt{AC\textsuperscript{2}MPC}.

\begin{equation} \label{eq:fu}
    x_{i+1} = f_{actual}(x_i, u_i) = f(x_i, u_i) + f_{u}(x_i, u_i)
\end{equation}

% \subsubsection*{\textbf{AC\textsuperscript{2}MPC}}
In \texttt{AC\textsuperscript{2}MPC}, an MPC controller and a learnt controller each generate a control input at control frequency $f_c$ during training and deployment. The observation space for training this controller is updated to additionally include the MPC's control input history and speed tracking history over the chosen history span $h_a$.

The chosen values for the action bounds, size of observation spaces and history length are enumerated in Table \ref{tab:controlParams}. The reward function is modified from the baseline learning controller to be $r_2(i)$ to additionally penalize control actions from compensatory controller that are unnecessary at certain operation regions, especially the regions where MPC can perform well. 
\begin{subequations} \label{eq:r2}
    \begin{align}
        r_2(i) & = W_{21} \cdot \frac{1}{1+ \left| v_{err} \right| } - W_{22} \cdot \frac{\sigma_{a_{rl}}}{N_{\sigma}} - W_{23} \cdot p_{23} \\
            & \text{s.t. } p_{23}=\begin{cases}
            1, & \text{if $a_{rl} > 0 \ \& \ v < v_{threshold}$}\\
            0, & \text{otherwise}.
          \end{cases}
    \end{align}
\end{subequations}
This also discourages the agent to violate the control bounds corresponding to actuator and vehicle limits. Further, the smoothness penalty is weighed lower because the MPC primary control inputs tend to be smooth already.  Since the model-based controller is expected to track small reference velocities well, $p_{23}$ is used to discourage unnecessary agent compensation at low speeds.
It must be noted that the primary controller generates a control input unaware of the compensatory controller. The compensatory controller inputs can thus be thought of as an external unobserved disturbance for the MPC that it may adjust to, in order to perform speed tracking. The control input passed to the system thus follows the policy $\Pi_{i}$.

\begin{equation} \label{eq:totalInput}
    \Pi_{i} = \Pi_{MPC,i} + \Pi_{\theta,i}
\end{equation}

The algorithmic description for this controller is given in Algorithm \ref{alg:mpcrl}. Here, $\mathcal{A}$ is the advantage function and $\epsilon$ is clip fraction used for the policy gradient updates. 
\begin{algorithm}[h]
\caption{\texttt{AC\textsuperscript{2}MPC}: Parallel compensation controller}\label{alg:mpcrl}
\begin{algorithmic}
\STATE 
% \STATE {\textsc{Get Feedback}}$(\mathbf{x}, u_{mpc,i-1}, \hat{u}_{mpc,i})$

\COMMENT{Executed at each control step}
\STATE $ \text{Solve MPC}  $     
\STATE      \hspace{0.5cm} $u_{mpc, i}, \hat{u}_{mpc, i+k} \gets \pi_{mpc}(\rho_{i},\mu_i, x_{i}) $
\STATE $\text{Get agent Action}$ 
\STATE  \hspace{0.5cm} $u_{rl, i} \gets \pi_{\theta}({x_{i}}_a, u_{mpc, i-k})$
\STATE      $\text{Compensate}$ 
\STATE \hspace{0.5cm} $u_{i} \gets u_{mpc, i}+ u_{rl, i} $
\STATE $\text{Simulate environment for 33 timesteps with ZOH}$ 
\STATE  \hspace{0.5cm} $x_{actual,i+1} \gets f_{actual}(x_i,u_i,T_s) $
\STATE $\text{Generate observations and feedback}$ 
\STATE  \hspace{0.5cm} $x_{actual,i+1}, s_{obs}  $
\STATE     $\text{If Training AC(PPO):}$ 

      % \hspace{0.5cm} \If{(if sample $\geq n_{steps}$):}
      \hspace{0.5cm} (if sample $\geq n_{steps}$):
      {      
      
        \hspace{0.9cm} $\theta_{i+1} \gets \arg\max_{\theta} E[\mathcal{L}(x_i, u_{rl}, \theta_i, \theta)] $
        
        \hspace{0.9cm} $\mathcal{L} = \min( \frac{\pi_{\theta}}{\pi_{\theta_{i}}} \cdot \mathcal{A}_i(x,u_{rl}), (1 \pm \epsilon) \cdot \mathcal{A}_i(x,u_{rl}))  $ 

        }           
\STATE \hspace{0.5cm}\textbf{return}  Null
\end{algorithmic}
\end{algorithm}

\section{Simulation Modeling and Verification}
Verification is carried out for the scenarios listed in Table \ref{tab:scenarioDescription}. The discussion of the simulation environment and each controller's tracking performance, smoothness and data efficiency is presented in this section. The scope of this study is to learn and analyze longitudinal tracking performance.
\begin{table}[h]
    \centering
    \caption{Description of test and validation simulation scenarios and terrain properties} % \resizebox{\columnwidth}{!}{}
    \begin{tabular}{rc|cc}
        \toprule
    	 Terrain Style & Terrain	\#  &	Scenario \# 	&	Reference	\\
          	  &  &	 	& Velocity	\\

        %
        % \midrule
        \midrule
        \midrule
         Loose deformable sand & Terrain-1  &	1A & constant \\
        % \midrule									
         & &	1B & varying \\
        \midrule									
        Sand over rocky terrain  & Terrain-2	&	2A & constant \\
        with high stiffness & &	2B &  varying \\
        \midrule									
        Clay-like and very soft & Terrain-3	&	3A &  constant \\
                % \midrule									
        & 	&	3B & varying \\
        \bottomrule
        \\
    \end{tabular}
    \label{tab:scenarioDescription}
    \\
    \begin{tabular}{c|ccc}
        \toprule
        Parameter & Terrain-1  & Terrain-2  & Terrain-3 \\
        \midrule
        \midrule
        Mohr's friciton limit [$^{\circ}$] & $30$ & $20$ & $14$ \\
        Soil stiffness, $k_{\phi}$ [-] &  $2 \cdot e^6$ &  $1 \cdot e^6$ &  $5 \cdot e^5$\\
        Cohesiveness coefficient, $k_c$ [-] & 0 & $1\cdot e^2$ & $1\cdot e^5$ \\
        Elastic stiffness before yield [Pa/m] & $2 \cdot e^8$ & $3 \cdot e^8$ & $2 \cdot e^7$  \\
        Shear coefficient [m] & $0.01$ & $0.005$ & $0.02$ \\
        Bekker exponent, n [-]   & $1.1$ & $1.0$ & $0.7$ \\
        Damping [Pa-s/m] & $3 \cdot e^4$ & $3 \cdot e^4$ & $5 \cdot e^4$ \\
        \bottomrule
    \end{tabular}
\end{table}
\subsection{High Fidelity Modeling for Off-road Driving}
\texttt{Project Chrono} \cite{Chrono2016} is used as the simulation environment for training, deployment and evaluation of the controllers. The \texttt{gym-chrono} \cite{gymchrono} custom environment wrapper that uses the \texttt{gymnasium} \cite{gym} library is leveraged to train the agent and evaluate it for different scenarios. Two terrain types are utilized in this study - rigid and deformable (an empirical Bekker-Wong model of the soil). The physics of the tire and deformable soil contact model is given in Eq. \eqref{eq:scm}.
\begin{equation} \label{eq:scm}
    \sigma = (\frac{k_c}{b} + k_{\phi}) \cdot y^n
\end{equation}
where, $\sigma$ is the contact pressure, $y$ is the wheel sinkage, $b$ represents the contact geometry and $k_c$, $k_{\phi}$ represent cohesiveness and stiffness of the soil. 
For deformable terrain-1, the parameters from default SCM terrain in the simulator were used.
For deformable terrain-2 model, these values are chosen to represent a harder soil with granularity. and for deformable terrain-3 model, these values were in the lower range to represent clay like cohesive soil. These values are enumerated in Table \ref{tab:scenarioDescription}.
The deformable terrain's mesh discretization size was chosen to \unit[$0.05$] {m} to sufficiently capture the interaction with the tire contact patch.

The vehicle used for this study is an off-road military style vehicle available in the simulator library, and is visualized in Fig.~\ref{fig:architecture}. This style of a vehicle is expected to run at moderately high speeds and is modelled with sufficiently high fidelity dynamics, powertrain, and driveline representations to fit the objective of this study. We assume ideal sensing setup for this study and obtain the vehicle states through simulation body frame's poses and velocities.
Since the computation times for a deformable terrain and a finite element deformable tire interaction yielded unrealistic training times, this study assumed tire rigidity when on deformable terrain.

\subsection{Simulation and Training Details}
To allow stability of contact force modelling and calculations, a sufficiently low simulation time step $T_s$ is chosen to be $\unit[3 \cdot e^{-3}$] {sec}. Since the control frequency is much lower, there are $33$ simulation steps between each control generation. To address this, a zero order hold is used on the control inputs from the constituting controllers across these time-steps during training and deployment.

To evaluate the performance and scalability of the different controllers with limited training, the agents were trained exclusively on constant speed scenarios, and only Terrain-1 was used during the training phase in this study.

\section{Results and Analysis}

\begin{figure}[h]
    \centering
    \includegraphics[width=\linewidth]{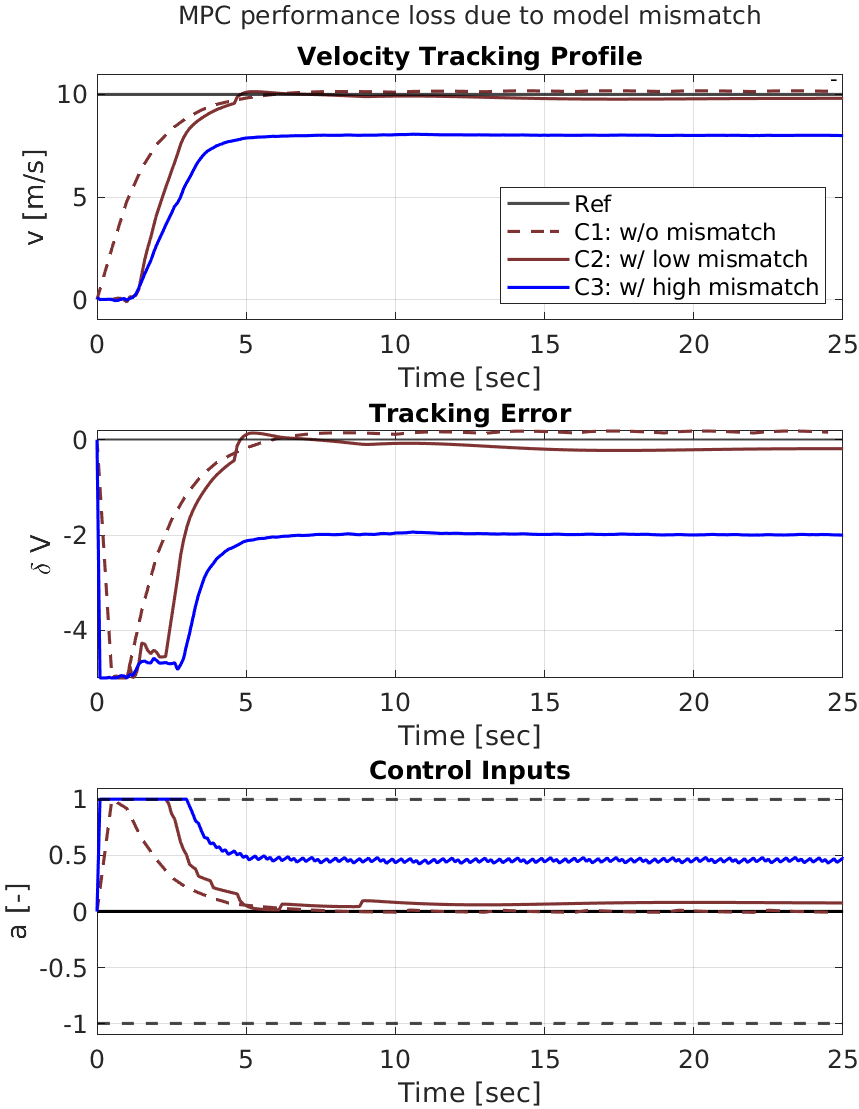}
    \caption{Need for learning augmentation to MPC: Even though control inputs are not restricted by the control bounds, vanilla MPC performs significantly worse on deformable terrain than on rigid terrain.}
    \label{fig:rigid-def}
\end{figure}

% \subsection{MPC's performance without augmentation} \label{sec:res:need}
A motivating example is shown in Fig.~\ref{fig:rigid-def} to emphasize the need for augmentation framework. Here, a model predictive controller is used to carry out the same task in three test cases. 
The baseline case C1 represents an on-road driving with no model mismatch between the controller and the simulator. The terrain is assumed rigid and the vehicle is modelled in the \texttt{acados} simulation as a kinematic one.
In case C2, a model-mismatch is introduced by using a higher fidelity vehicle model using the chrono simulator with rigid terrain. For this rigid terrain and TMeasy \cite{rill2020road} tire interaction, a friction coefficient of $0.9$ is selected. 
Case C3 additionally sees a deformable terrain in the simulation. It is clear that this model-based controller (formulated as in Section \ref{sec:mpc}) fails to perform as is and needs augmentation.

Next, we discuss how the compensation framework helps to improve longitudinal performance in the six scenarios listed above.
For each of these scenarios, we plot and and discuss the performance of the MPC, AC and \texttt{AC\textsuperscript{2}MPC} controller for the same task with the same random seeds.

\subsection{Comparison for Terrain-1 (Loose, granular and deformable soil): Scenarios 1A and 1B}
\begin{figure}[]
    \centering
    \begin{subfigure}{\linewidth}
       \includegraphics[width=\linewidth]{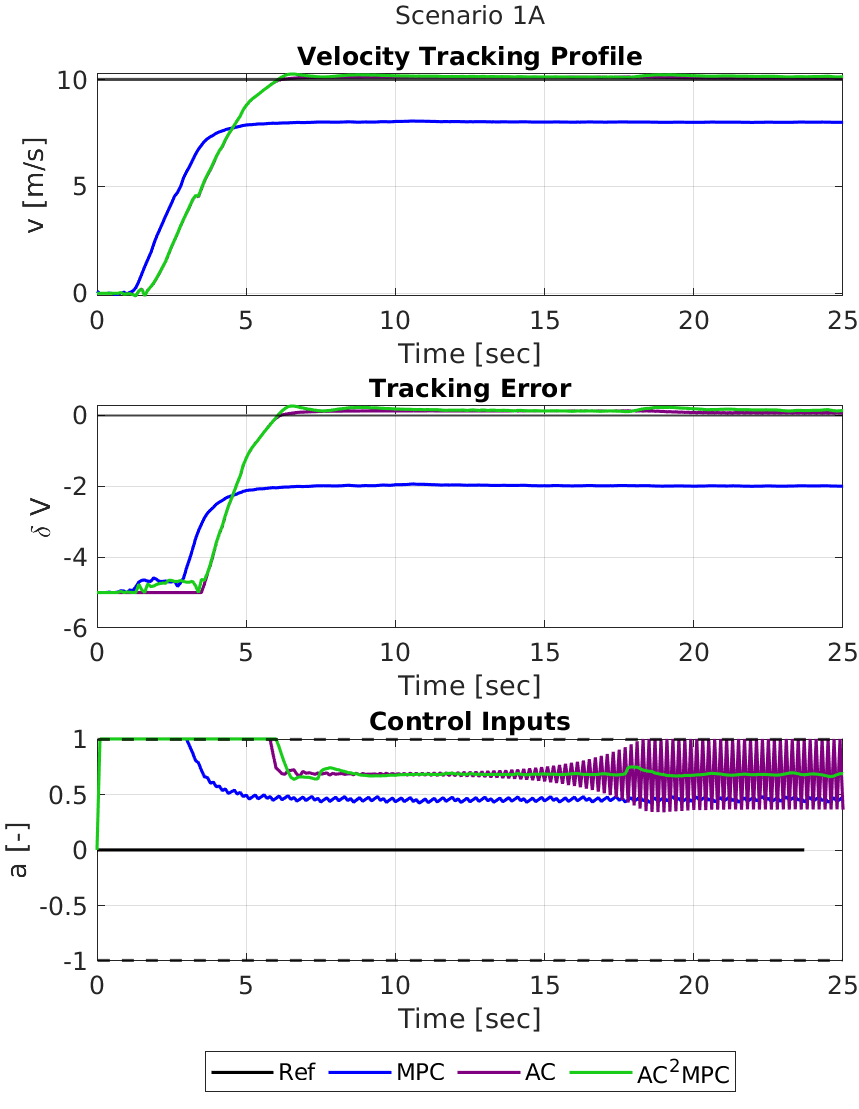}
        % \caption{Tracking performance and control efforts for scenario 1: constant speed tracking on terrain-1} 
    \end{subfigure}
        \begin{subfigure}{\linewidth}
       \includegraphics[width=\linewidth]{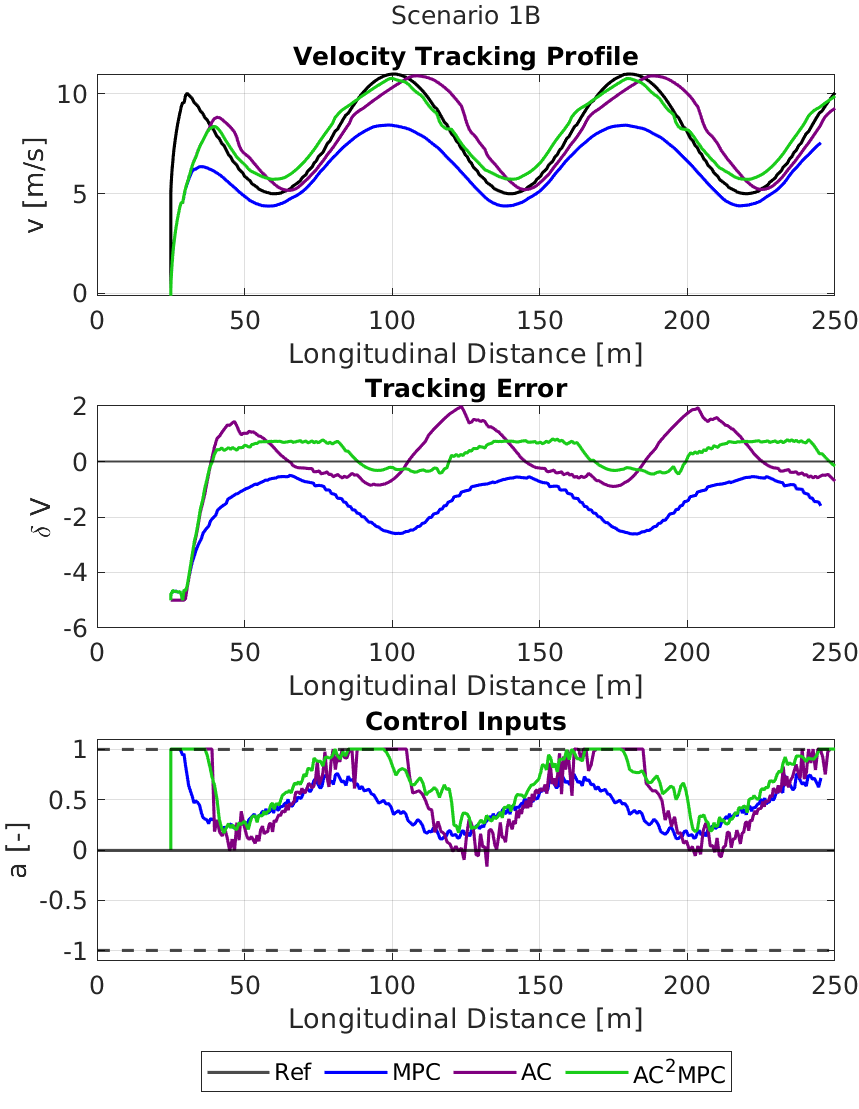}
        % \caption{Tracking performance and control efforts for scenario 2: varying speeds tracking on terrain-1.} 
    \end{subfigure}
    \caption{Comparison for scenarios 1A and 1B: velocity tracking on terrain-1 for constant and varying reference velocities respectively.}
    \label{fig:s1AB}
\end{figure}

The scenarios 1A and 1B corresponding to constant and varying reference velocity respectively. The tracking performance of each controller is plotted in Fig.~\ref{fig:s1AB}. 
This is the case where the evaluation terrain is the same as the one exposed to the agent during training. This terrain has the properties that are closer to a granular, sandy and deformable soil with moderate Mohr's friction limit. 
For 1A, it is clear that the MPC under-performs with abundantly available unexploited control space due the reasons explained in the previous section.
The RL controller learns to behave like a PID controller upon training convergence. It is able to track the reference velocity profile on the given surface well, but the control inputs become very jittery as this target is maintained as shown in the third plot for scenario 1A in Fig.~\ref{fig:s1AB}. This is owed to its inability to discriminate between a same high valued scalar multi-objective reward from input trajectories exhibiting high and low jerk values.
The compensated learning controller \texttt{AC\textsuperscript{2}MPC} is able to track the target velocity well, with its control input being the smoothest and most appropriate for implementation on a real vehicle platform. The rms tracking errors in velocity, denoted by $\delta V_{rms}$ and rms jerk values along the simulation are recorded in Table \ref{tab:results}. The lowest RMS errors among all controller performances for one complete simulation are provided in bold text.

This $28.3\%$ gain in performance (based on $\delta V_{rms}$) and the difference in control inputs between the MPC and compensated controller arise as a result of the learnt compensation input that is added to the system. It should be noted that this added control input is lesser than that of the RL controller and much smoother. The average rms error of our controller  is lowest. 

For scenario 1B, which corresponds to tracking a varying velocity profile previously unseen by the agents, a similar performance and jerk trends follow, and it is seen that the tracking error for our controller is the lowest with $19.6\%$ lower $\delta V_{rms}$ than MPC. Our controller also generalizes better than the AC controller for a new velocity profile, with $7.67\%$ lower $\delta V_{rms}$. It is noted that a position based analysis is provided for 1B because this study provides position based velocity references on the off-road terrain. This aligns with the approach of deciding reference velocities of the given vehicle based of traversibility analyses. We also notice that the performance gain over AC controller is higher in the varying velocity case. This is expected becasue the agent was only trained over constant reference velocity scenarios.
For practicality, all these controller commands are saturated to emulate actuator limits before being passed into the simulator for training and deployment.

Another thing to note is the dead zone in first plot of Fig.~\ref{fig:s1AB} from $\unit[0 \text{ to } 1.5 $] {sec}. Here, there there exist non-zero control inputs, but no velocity change for the vehicle in simulation. This duration represents the time it takes for the vehicle to settle down on the soil and overcome the frictional resistance and drag due to the tire and deformable terrain interaction. This also constitutes the stabilization of the simulation contact forces between the vehicle tires and the terrain. Once this settle time is over, the vehicle starts to move forward in simulation.

\subsection{Comparison for Terrain-2 (Sandy with Higher Stiffness): Scenarios 2A and 2B}
\begin{figure}[]
    \centering
    \begin{subfigure}{\linewidth}
    \includegraphics[width=\linewidth]{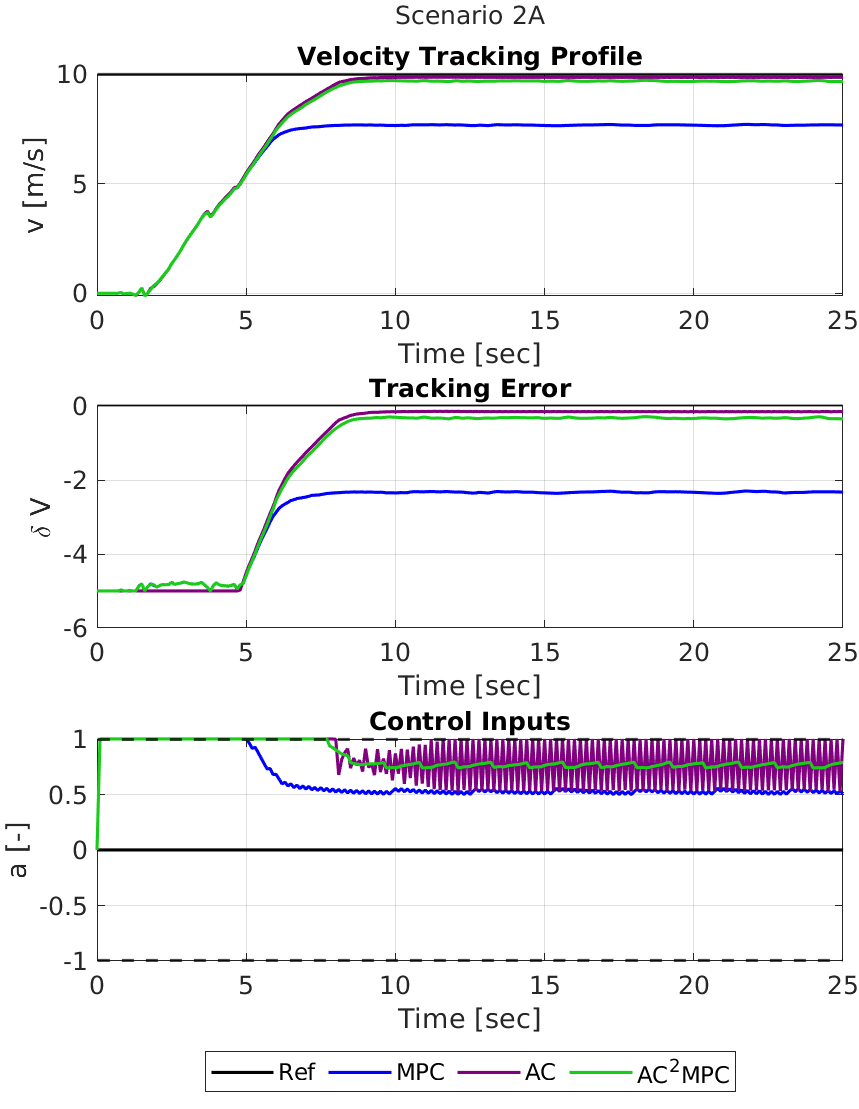}
    \end{subfigure}
        \begin{subfigure}{\linewidth}
    \includegraphics[width=\linewidth]{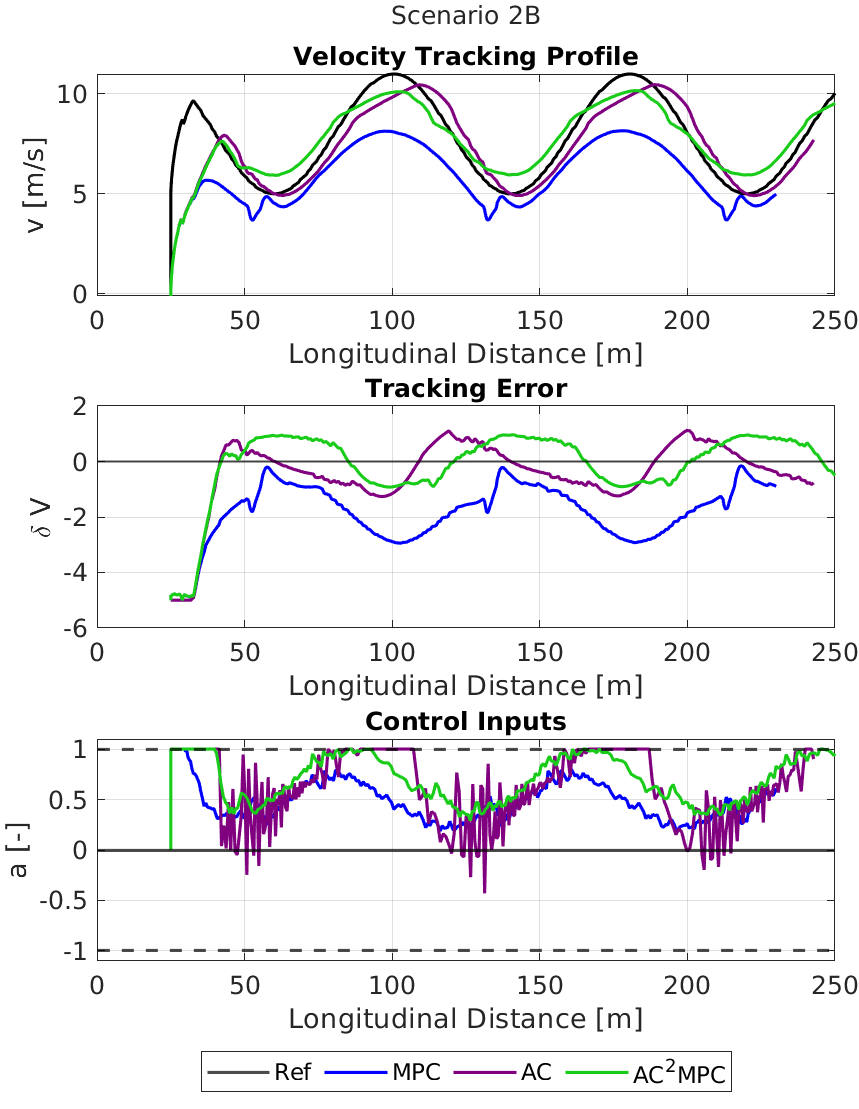}
    \end{subfigure}
    \caption{Comparison for scenarios 2A and 2B: velocity tracking on terrain-2 for constant and varying reference velocities respectively.}
    \label{fig:s2AB}
\end{figure}

For scenarios 2A and 2B over terrain-2, the tracking errors and control inputs are plotted in Fig.~\ref{fig:s2AB}. This terrain corresponds to a sandy rough road, with little deformability in the soil. A Mohr's friction limit angle of $20^{\circ}$ represents a granular soil and a higher contact, and elastic stiffness lend to a harder soil due to the slight rocky and uneven terrain. This was chosen to evaluate the behavior of the controllers on a soil that is closer to rigid terrain than the one the agents are trained on, and is not exposed to the agent during training.

For constant reference tracking over terrain-2, it is seen that the standalone controllers are not able to track the profile well, despite available control space. Our controller outperforms MPC by $25.6\%$ and AC by $0.9\%$. The first fails to account for the the granularity of the terrain and the latter is unable to generalize what it learnt on terrain-1, whereas \texttt{AC\textsuperscript{2}MPC} generalizes much better and maintains high smoothness of control inputs with an rms jerk of $\unit[1.21]{m/s^3}$. The rms jerk value for the standalone learnt controller is $\unit[3.94]{m/s^3}$ and follows a bang-bang like control.

In case of varying references, MPC under-performs in the same fashion as in scenario 2A. We also notice that the AC controller tends to lag behind the reference profile. This is explained by the fact that its policy was trained under limited constant reference scenarios and thus generalizes poorly to varying profiles. Our controller outperforms the optimal controller by $20.3\%$ and the learning based controller by $0.8\%$ in tracking.
In both 2A and 2B, the AC controller is highly jerky when it operates away from the upper control bounds and is unfit to deploy on a real vehicle. Our controller is able to track crests of the profile much better than the troughs because the model-based component still tends to under-estimates the granularity model mismatch. However, the control effort is much smoother than standalone AC with $52\%$ decrease in rms jerk along the trajectory.

% Scene: 0, A, 2.452, 1.801,1.757,   2.835,4.593,1.014,    28.340, 2.43, 
% Scene: 1, A, 2.140, 1.862,1.720,   3.093,1.273,1.098,    19.633, 7.67, 
% Scene: 2, B3, 3.020, 2.268,2.247,   2.996,3.947,1.218,   25.608, 0.92, 
% Scene: 3, B3, 2.497, 2.004,1.988,   3.049,2.470,1.101,   20.389, 0.78,
% Scene: 2, C2, 2.492, 2.198,2.157,   2.918,4.662,1.285,    13.444, 1.89, 
% Scene: 3, C2, 1.998, 1.901,1.933,   3.099,2.546,1.120,   3.278, -1.66, 

\subsection{Comparison for Terrain-3 (Clay-like Soft Soil): Scenarios 3A and 3B}
Scenarios 3A and 3B, utilize terrain-3 which is modelled to imitate a soft and cohesive clay soil with low stiffness and friction limit angle. This is the case where the model mismatch between the vehicle-terrain interaction and the simulated system is higher than that seen during training. 

\begin{figure}[]
    \centering
    \begin{subfigure}{\linewidth}
    \includegraphics[width=\linewidth]{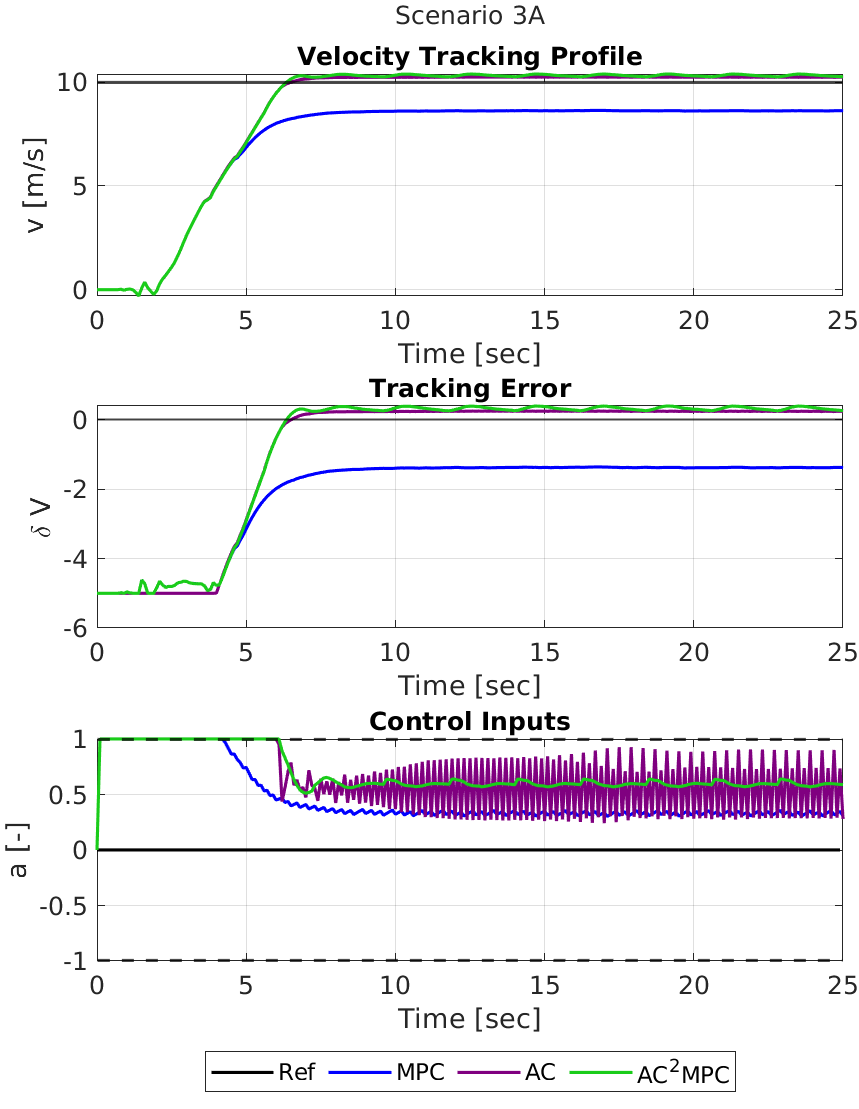}
    \end{subfigure}
    \begin{subfigure}{\linewidth}
    \includegraphics[width=\linewidth]{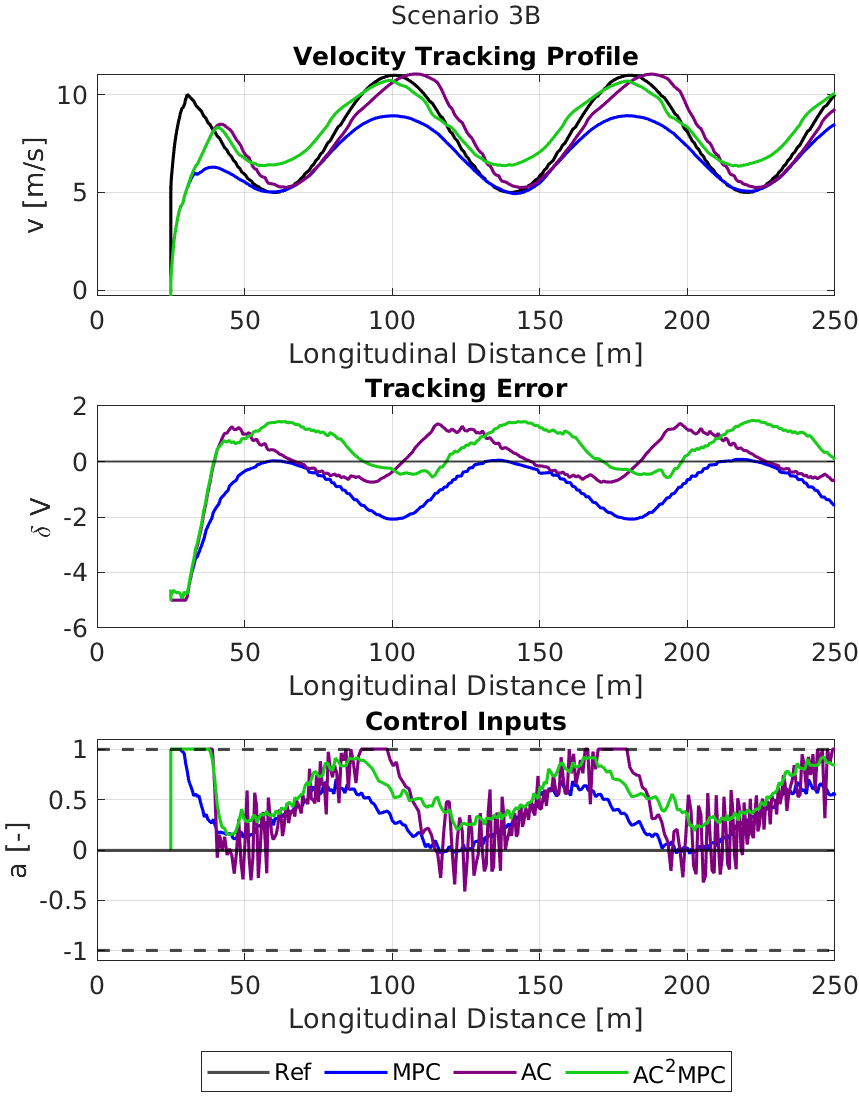}
    \end{subfigure}
    \caption{Comparison for scenarios 3A and 3B: velocity tracking on terrain-3 for constant and varying reference velocities respectively.}
    \label{fig:s3AB}
\end{figure}

As shown in Fig.~\ref{fig:s3AB}, it is obvious that the MPC severely under-performs due to the model mismatch. For 3A, the our controller outperforms the standalone controllers by $13.4\%$ and $1.9\%$ in terms of tracking errors. The controllers have a significantly higher steady state control input in this case, because it takes more acceleration to overcome the soil cohesion and deformability. It must be noted that MPC, missing the actual model information starts approaching the steady state much sooner to track the same reference velocity.

In case of 3B, we notice that \texttt{AC\textsuperscript{2}MPC} outperforms MPC by $3.3\%$ to minimize tracking errors, and trails behind the AC controller by $-1.6\%$. This performance loss is small, with the difference in the  $\delta V_{rms}$ being $\unit[0.03]{m/s}$. 
There is also a lag because the latter does not have a preview of the upcoming velocity changes and is unable to account for the dynamic lag arising from softness of the soil. The performance loss in This causes the learning controller's inputs to ride the input constraints in order to make up for the tracking lag. Despite this constraint riding behavior, the rms jerk for learning controller is higher than our controller's by $50\%$. 
\begin{figure}[] 
    \centering
    \begin{subfigure}{\linewidth}
    \includegraphics[width=\linewidth]{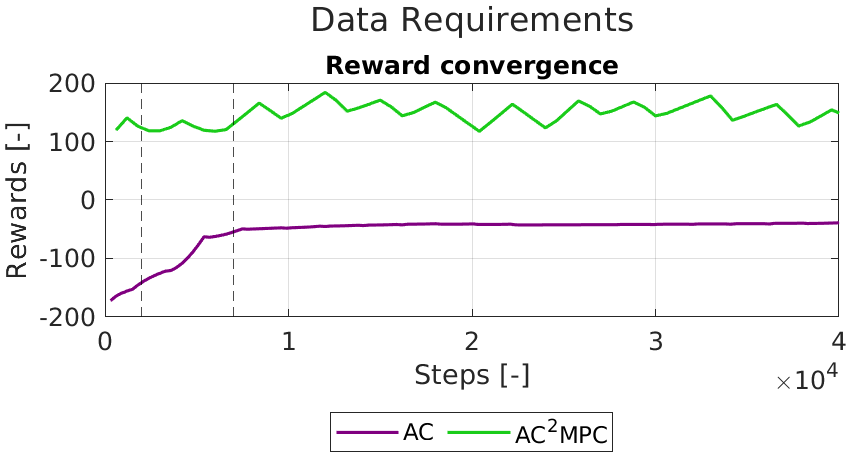}
    \caption{Rewards for AC and AC\textsuperscript{2}MPC are shown here. It is seen that AC\textsuperscript{2}MPC starts at a high reward value already and converges within 5,000 steps.}
    \end{subfigure}
    \begin{subfigure}{\linewidth}
    \includegraphics[width=\linewidth]{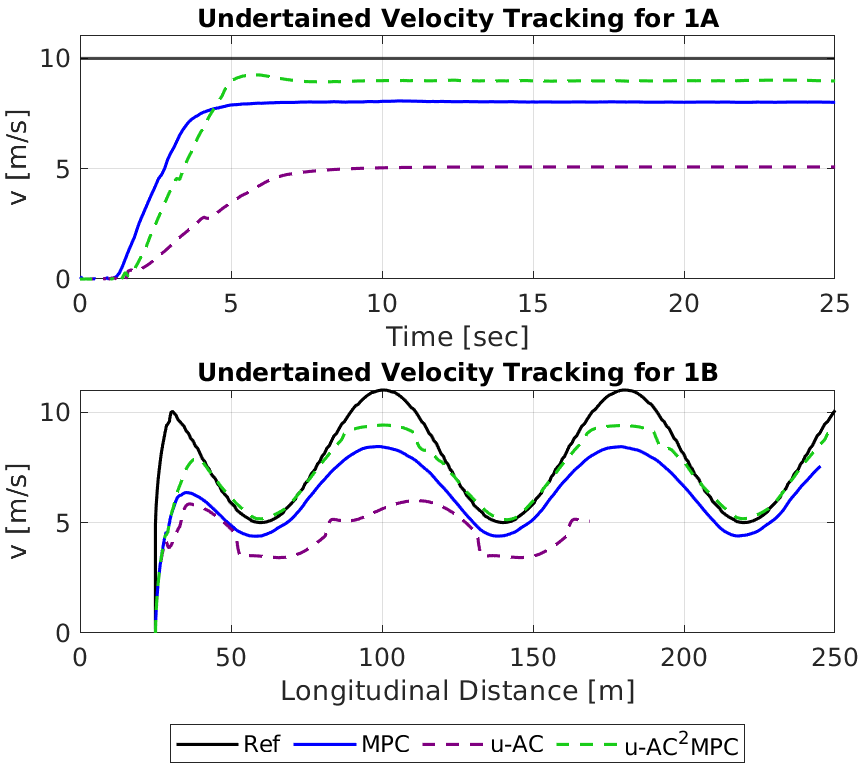}
    \caption{At only 2,000 steps of training, the performance of under-trained AC (u-AC) is much worse than under-trained AC\textsuperscript{2}MPC (u-AC\textsuperscript{2}MPC) for both constant and varying velocity tracking scenarios.}
    \end{subfigure}
    \caption{Data requirements for compensated controller framework are lower.}
    \label{fig:data}
    
\end{figure}
\begin{table}[]
    \centering
    \caption{Statistics for all scenarios and controllers: RMS tracking error and RMS jerk values for one complete simulation.} % \resizebox{\columnwidth}{!}{}
    \begin{tabular}{r|l|ccc}
        \toprule
    	 Scenario & Controller	&	$\delta V_{rms}$ [m/s]  &	$RMS Jerk [m/s^3]$	\\
      % Scenario & Controller	&	RMS Error	&	Smootheness	&	Data Reqd.	\\
        %
        \midrule	
        \midrule									

        Scenario 1A & MPC	&	2.452  &	2.835	 \\
             & AC	&	1.801 & 4.593\\
             & AC\textsuperscript{2}MPC	&	\textbf{1.757} & \textbf{1.014} \\
             & u-AC	&	4.948 & 0.47\\
             & u-AC\textsuperscript{2}MPC	& 2.019 & 1.04	 \\
         % & AC\textsuperscript{2}MPC-C	&	\unit[12.95]{m/s}	& \unit[41.0]{\%} &	\unit[210.4]{s}	\\
        %
        \midrule									
        Scenario 1B & MPC	&	2.140  &	3.093	 \\
             & AC	&	1.862 & 1.273 \\
             & AC\textsuperscript{2}MPC	&	\textbf{1.720} & \textbf{1.098} \\
             & u-AC	&	3.314 & 0.47\\
             & u-AC\textsuperscript{2}MPC	& 1.666 & 0.86	 \\
        \midrule			% using terrB3		
 
        Scenario 2A & MPC	&	3.020  & 2.996 \\
             & AC	&	2.268 & 3.947\\
             & AC\textsuperscript{2}MPC	&	\textbf{2.247} & \textbf{1.218} \\
        \midrule									
        Scenario 2B & MPC	&	2.497  & 3.049 \\
             & AC	&	2.004 & 2.470 \\
             & AC\textsuperscript{2}MPC	&	\textbf{1.988} & \textbf{1.101} \\       
        \midrule							

        Scenario 3A & MPC	&	2.492  & 2.918	 \\
             & AC	&	2.198 & 4.662 \\
             & AC\textsuperscript{2}MPC	&	\textbf{2.157} & \textbf{1.285} \\
        \midrule									
        Scenario 3B & MPC	&	1.998  & 3.099\\
             & AC	&	\textbf{1.901}& 2.546 \\
             & AC\textsuperscript{2}MPC	&	\textbf{\textit{1.933}} & \textbf{1.120} \\            
             
        \bottomrule
    \end{tabular}
    \label{tab:results}
\end{table}

\subsection{Data Efficiency Analysis}
Fig. \ref{fig:data}(a) shows that the data requirement for training \texttt{AC\textsuperscript{2}MPC} controller is much lower. Much like in RPL \cite{silver2018residual}, this is because the compensated control policy requires less training to learn the compensation as compared to learning the entire control policy from scratch. 
The rewards start off high as the compensated controller training begins and converge at less than $5,000$ steps of training for the compensated controller. 
In contrast, it takes around $20,000$ steps for standalone reinforcement learning based training to converge. Even after convergence, the tracking performance of our controller is superior to learning based controller for almost all the chosen evaluation scenarios.

Moreover, in a real world controller deployment for off-road driving applications, training data available to the agent is generally uncertain and, to some extent under-representative of the scenarios to be handled. So, we are interested in understanding how our controller performs when under-trained. 
To evaluate controller's fitness after a limited simulation training, we extract control policies from AC and compensated controllers at only $2,000$ steps of training. These under-trained policies, denoted as \texttt{u-AC} and \texttt{u-AC\textsuperscript{2}MPC} here, are used to re-evaluate performance for scenarios 1A and 1B as shown in Fig.~\ref{fig:data}(b). In both the constant and varying reference cases, all controllers fail to reach the reference velocity range. Since our controller is designed to focus on the unmodeled dynamics, it is seen to already improve upon the model based controller here.

In terms of tracking error, our under-trained controller still outperforms both MPC and \texttt{u-AC} controllers by $14.8\%$ and $59.1\%$ respectively. This lends more trust in the compensated controller for improved and safer performance when deployed for on-policy learning on the real platforms. This can be especially important because learning based controller usually suffer from simulation to reality transfer gaps.

\subsection{Comparison to other hybrid approaches}
\begin{figure}[h]
    \centering
    \includegraphics[width=\linewidth]{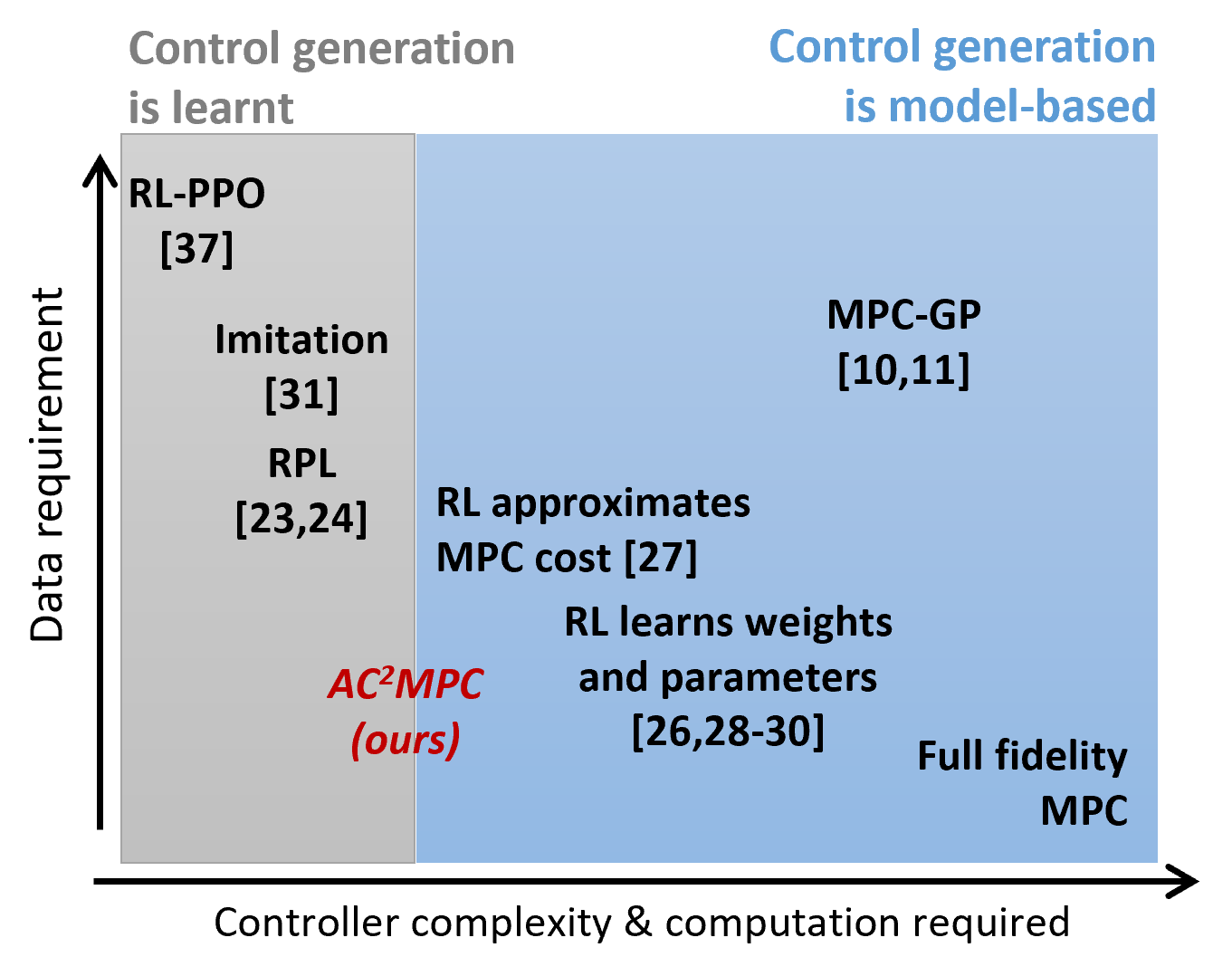}
    \caption{Our approach, AC\textsuperscript{2}MPC can perform similar or better than standalone controllers while requiring lesser data and a moderately simple model for controller formulation.}
    \label{fig:litViz}
\end{figure}

Other hybrid approaches \cite{silver2018residual} -\cite{delft} mentioned in section \ref{sec:relevant} that exploit RL and MPC paradigms are all sequential and/or frequency stepped by design and thus a one-to-one comparison to our framework is not feasible. However, all approaches that lend to control generation by the trained agent such as in RPL \cite{silver2018residual}, have higher data requirements compared to our approach. Further, do not promise safe actions for unseen scenarios. In the approaches where parameters or costs of the optimal controllers are learnt, the optimization and/or formulation are intensive and need higher computation as in \cite{zarrouki2024safe}. This is represented in Fig.~\ref{fig:litViz}, where MPC-GP \cite{fisac2018general} on the top right corner requires a complex enough acquisition function and lots of data to constantly update the confidence bounds for the dynamics. A full fidelity MPC needs no data but can be difficult to formulate and solve in real time. Our approach lies in the region that utilizes moderately low controller complexity and very less data to perform as well or better than the standalone controllers. And owing to a parallel approach, the optimal controller always attempts to ensure safety constraint satisfaction.

\section{Conclusion and Future Work}
In this study, we developed and evaluated a parallel controller framework for reinforcement learning compensated model predictive controller to track reference velocities on multiple unknown off-road deformable terrains. The controllers were evaluated using a high-fidelity simulator to capture the tire and deformable terrain interaction. Through evaluation over constant and varying velocity reference profiles, we found that for all the studied scenarios, \texttt{AC\textsuperscript{2}MPC} performs statistically better than the standalone MPC and RL controllers in terms of rms values of tracking error. We also analyzed the smoothness of these controllers using rms jerk metrics, their tracking lags and some constraint riding behavior and these statistical results are presented together in Table \ref{tab:results}. Data efficiency of our controller was quantified against the reinforcement learning controller and found to be better, even when the controller is under-trained.

A limitation of this controller is the inability of the constituent controllers to actively predict each other's intentions and share cooperative information. This improvement opportunity will be investigated as the next step. Since this evaluation was limited to only longitudinal performance, we plan to extend the framework to lateral tracking performance to understand its scalability to steering controls. The next step will verification of this controller on a Polaris RZR off-road drive-by-wire platform over terrain with unknown characteristics and benchmark the improvement over purely model-based and purely learning based controllers.

\section*{Acknowledgment}
This work was supported by Clemson University’s Virtual Prototyping of Autonomy Enabled Ground Systems (VIPR-GS), under Cooperative Agreement W56HZV-21-2-0001 with the US Army DEVCOM Ground Vehicle Systems Center (GVSC).

\bibliographystyle{IEEEtran}

\bibliography{bibliography}

\newpage

%%%%%% bios
\begin{IEEEbiography}
[{\includegraphics[width=1in, height=1.25in, clip, keepaspectratio]{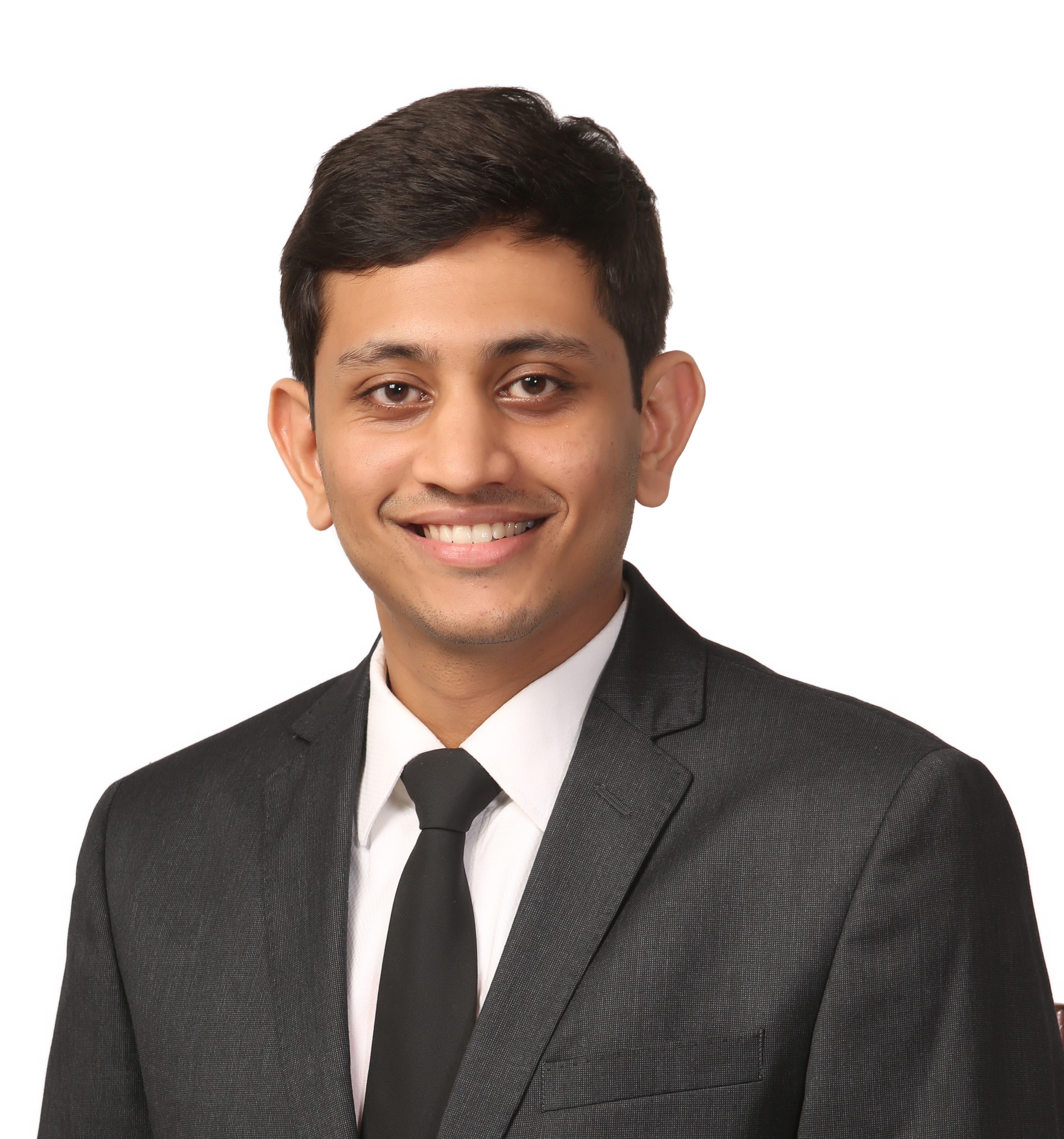}}]
{Prakhar Gupta}
Prakhar Gupta received his B.E. degree in Mechanical Engineering from Manipal University, India in 2017. He was a CAE Analyst at Mercedes Benz R\&D India from 2017-2020, developing vehicle dynamics multi-body simulations for heavy vehicles. He is currently a Ph.D. student at the Department of Automotive Engineering at Clemson University International Center for Automotive Research, USA. His research interests include learning augmentation to model-based optimal control techniques for autonomous vehicles and robots.
\end{IEEEbiography}

% \vfill

\begin{IEEEbiography}
[{\includegraphics[width=1in, height=1.25in, clip, keepaspectratio]{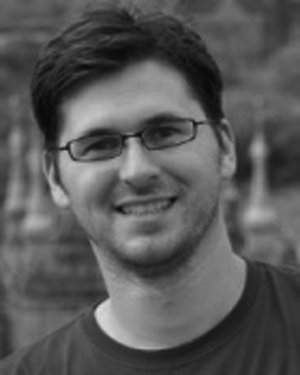}}]
{Jonathon M. Smereka}
Jonathon M. Smereka received the B.Sc. Eng. degree in computer engineer, electrical engineering, and engineering mathematics from the University of Michigan at Dearborn, Dearborn, MI, USA, in 2009, and the M.S. and Ph.D. degrees in computer and electrical engineering from Carnegie Mellon University, Pittsburgh, PA, USA, in 2016. He is currently the Ground Vehicle Robotics Senior Technical Expert for research with the U.S. Army DEVCOM Ground Vehicle Systems Center (GVSC, formerly TARDEC) in Warren, MI, USA. In this role, he is the Senior Technical Advisor for an organization focused on developing and delivering robotic and autonomous capabilities and test and evaluation procedures to Army acquisition programs. In 2007, he began his career with the U.S. Army TARDEC, as an Intern with the High Performance Computing Team, left for graduate school in 2010, and then rejoined again in 2016, through the DoD Science, Mathematics, and Research for Transformation Fellowship. He worked on research and development of artificial intelligence, machine learning, image and signal processing, and computer vision with respect to biometrics, automated target recognition, and ground vehicle autonomy.
\end{IEEEbiography}
% \newpage

\begin{IEEEbiography}
[{\includegraphics[width=1in, height=1.25in, clip, keepaspectratio]{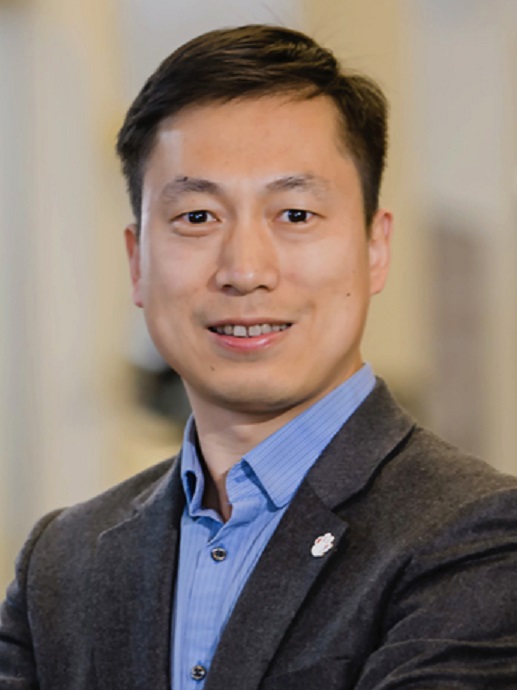}}]
{Yunyi Jia}
    Yunyi Jia (S`08-M`15-SM`20) received the B.Sc. degree from National University of Defense Technology,  M.Sc. degree from South China University of Technology and Ph.D. degree from Michigan State University. He is currently the McQueen Quattlebaum Associate Professor and the director of Collaborative Robotics and Automation (CRA) Lab in the Department of Automotive Engineering at Clemson University International Center for Automotive Research. His research interests include collaborative robotics, automated vehicles and advanced sensing systems.
\end{IEEEbiography}

\vfill

\end{document}